\def\year{2021}\relax
\documentclass[letterpaper]{article} 
\usepackage{aaai21}  
\usepackage{times}  
\usepackage{helvet}  
\usepackage{courier}  
\usepackage{url}  
\usepackage{graphicx}  
\usepackage{float}
\usepackage{multirow}
\usepackage{diagbox}
\usepackage{amsmath}
\usepackage{amsthm}
\usepackage[linesnumbered,titlenumbered,ruled,vlined,commentsnumbered,noend]{algorithm2e}

\usepackage{times}
\usepackage{xcolor}
\usepackage{soul}
\usepackage[utf8]{inputenc}
\usepackage[small]{caption}
\usepackage{comment}
\usepackage{color}

\usepackage{epsfig}
\usepackage{graphicx}
\usepackage{amsmath}
\usepackage{amssymb}
\usepackage{amsthm}
\usepackage{dsfont}
\usepackage{natbib}

\usepackage{url}
\usepackage{multirow}
\usepackage{enumitem}
\usepackage{moresize}
\usepackage{epstopdf}
\usepackage{array}

\usepackage[switch]{lineno}  %
\usepackage{booktabs}
\usepackage{pifont}
\usepackage{wrapfig}
\usepackage{subfigure}
\usepackage[hang,flushmargin]{footmisc} 

\usepackage{xspace}

\usepackage[colorlinks,
linkcolor=red,
citecolor=blue
]{hyperref}

\usepackage{xspace}
\makeatletter
\DeclareRobustCommand\onedot{\futurelet\@let@token\@onedot}
\def\@onedot{\ifx\@let@token.\else.\null\fi\xspace}
\def\eg{\emph{e.g}\onedot} 
\def\ie{\emph{i.e}\onedot} 
 
\def\etc{\emph{etc}\onedot}

\makeatother

\definecolor{darkorange}{rgb}{1.0, 0.55, 0.0}
\definecolor{lincolngreen}{rgb}{0.11, 0.35, 0.02}
\definecolor{cornflowerblue}{rgb}{0.39, 0.58, 0.93}
\definecolor{cobalt}{rgb}{0.0, 0.28, 0.67}

\usepackage{colortbl}
\definecolor{tabgray}{rgb}{0.85,0.85,0.85}
\definecolor{top1}{rgb}{1.0, 0.6, 0.6} 
\definecolor{top2}{rgb}{0.98, 0.91, 0.71}
\definecolor{top3}{rgb}{0.91, 1.0, 1.0}
\definecolor{top1-2}{rgb}{1.0, 0.66, 0.66} 
\definecolor{top1-3}{rgb}{1.0, 0.72, 0.72} 
\definecolor{top1-4}{rgb}{1.0, 0.78, 0.78} 
\definecolor{top1-5}{rgb}{1.0, 0.84, 0.84} 
\definecolor{top1-6}{rgb}{1.0, 0.90, 0.90} 
\definecolor{top1-7}{rgb}{1.0, 0.96, 0.96}

\title{EfficientDeRain: Learning Pixel-wise Dilation Filtering for High-Efficiency Single-Image Deraining}

\author {
    Qing Guo\textsuperscript{\rm 1}\textsuperscript{\rm $*$},
    Jingyang Sun\textsuperscript{\rm 2}\thanks{Qing Guo and Jingyang Sun are co-first authors and contribute equally to this work.},
    Felix Juefei-Xu\textsuperscript{\rm 3},
    Lei Ma\textsuperscript{\rm 2}\thanks{Lei Ma is the corresponding author: \href{mailto:malei@ait.kyushu-u.ac.jp}{malei@ait.kyushu-u.ac.jp}.},
    Xiaofei Xie\textsuperscript{\rm 1},
    Wei Feng\textsuperscript{\rm 4},
    Yang Liu\textsuperscript{\rm 1} \\
}

\affiliations {
    \textsuperscript{\rm 1} Nanyang Technological University, Singapore \\
    \textsuperscript{\rm 2} Kyushu University, Japan \\
    \textsuperscript{\rm 3} Alibaba Group, USA \\
    \textsuperscript{\rm 4} College of Intelligence and Computing, Tianjin University, China \\
}

\begin{document}
\maketitle

\begin{abstract}

Single-image deraining is rather challenging due to the unknown rain model.
Existing methods often make specific assumptions of the rain model, which can hardly cover many diverse circumstances in the real world, making them have to employ complex optimization or progressive refinement.
%
This, however, significantly affects these methods' efficiency and effectiveness for many efficiency-critical applications.
To fill this gap, in this paper, we regard the single-image deraining as a general image-enhancing problem and originally propose a model-free deraining method, \ie, \textit{EfficientDeRain}, which is able to process a rainy image within 10~ms (\ie, around 6~ms on average), over 80 times faster than the state-of-the-art method (\ie, RCDNet), while achieving similar de-rain effects.
%
We first propose the novel \textit{pixel-wise dilation filtering}. In particular, a rainy image is filtered with the pixel-wise kernels estimated from a \textit{kernel prediction network}, by which suitable multi-scale kernels for each pixel
can be efficiently predicted.
Then, to eliminate the gap between synthetic and real data, we further propose an effective data augmentation method (\ie, \textit{RainMix}) that helps to train network for real rainy image handling.
We perform comprehensive evaluation on both synthetic and real-world rainy datasets to demonstrate the effectiveness and efficiency of our method.
We release the \textit{EfficientDeRain} in \url{https://github.com/tsingqguo/efficientderain.git}.
\end{abstract}

\section{Introduction}\label{sec:intro}

Rain patterns or streaks captured by outdoor vision systems (\eg, stationary image or dynamic video sequence), often lead to sharp intensity fluctuations in images or videos, causing performance degradation for the visual perception systems \cite{garg2005does,garg2007vision} across different tasks, \eg, pedestrian detection \cite{Wang2018cvpr},  object tracking \cite{Li18}, semantic segmentation \cite{MaCoSNet}, \etc.
%
%
As a common real-world phenomenon, it is almost mandatory that an all-weather vision system is equipped with the deraining capability for usage. A deraining method processes the rain-corrupted image/video data and removes the rain streaks, with the intention to achieve good image quality for the downstream vision tasks. 
%
%

In many real-time applications that are efficiency-sensitive and critical, (\eg, vision-based autonomous driving or navigation), being able to perform deraining \emph{efficiently} on-chip is of great importance. A deraining algorithm achieving both high efficiency and high performance (\eg, in terms of PSNR, SSIM), while remaining at low overhead, is of great importance for practical usage.

\begin{figure}[t]
\centering
\includegraphics[width=1\columnwidth]{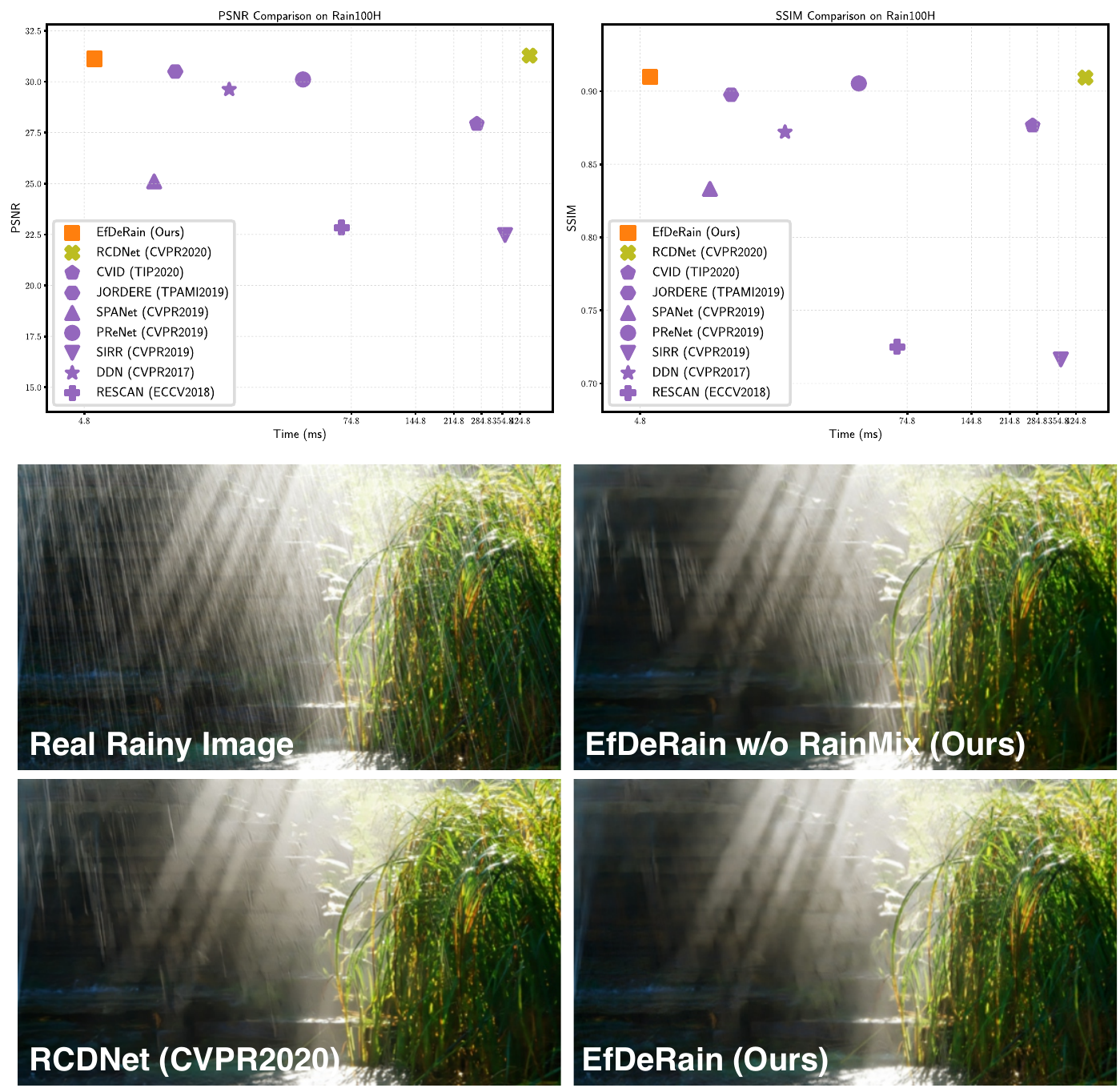}
\caption{Top: Comparison results (\ie, PSNR vs. Time and SSIM vs. Time) on the challenging Rain100H dataset. Down: an example of using EfDeRain, EfDeRain without RainMix and the state-of-the-art RCDNet \cite{rcd} to handle a real rainy image. Note that, the time cost of all compared methods are one-by-one evaluated on the same PC.}
\label{fig:teaser}
\end{figure}
\begin{figure*}
\centering
\includegraphics[width=0.8\linewidth]{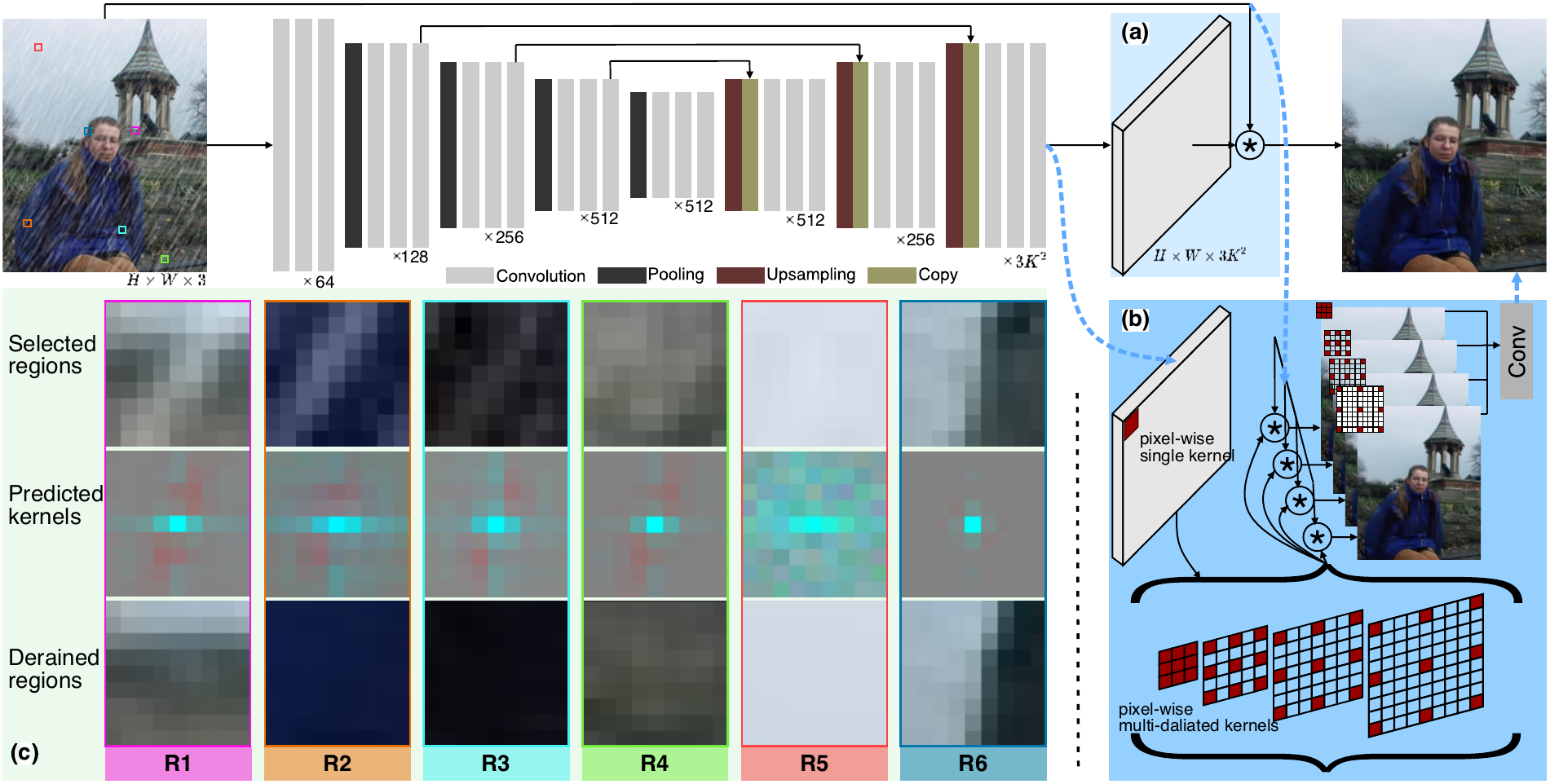}
\caption{Pipeline of our \textit{EfficientDeRain}. (a) represents the pixel-wise image filtering introduced in Sec.~\ref{subsec:kernel_pred_} where each pixel is processed by an exclusive kernel predicted by a kernel prediction network. (b) we further extend this simple structure and propose the pixel-wise dilation filtering in Sec.~\ref{subsec:multi_dilated_kernel pred} to handle multi-scale rain streaks. (c) we show that the predicted kernels can be adapted to different image contents while recovering the object boundary in the original image.}
\label{fig:pipeline}
\end{figure*}

Although we have witnessed the recent progress for deraining,  
existing methods mostly focus on studying the physical models of rain and background layers, removing rain streaks via solving an optimization problem and employing the power of deep learning and some priors, \eg, the rain having similar local patterns across the image domain \cite{rcd,li2016cvpr,wei2019cvpr,hu2019cvpr,jorder}.
An essential problem, \ie, the efficiency of the derain method, however, is largely neglected so far, which brings great limitation for real-time applications.

%
In particular, existing methods (see Sec.~\ref{sec:related}) often heavily rely on various rain-generating assumptions and rain-background models, whose goal is to revert such a rain-adding step in the deraining process, involving heavy iterative optimization and subsequent refinement steps.
However, some caveats of theirs cannot be overlooked: \ding{182} the rain model assumptions adopted by many of the algorithms may be limited, which do not well represent and reflect the real-world rain patterns. The models based on these rain model assumptions may not perform as strongly under the real-world scenarios; \ding{183} many of the existing methods are computationally expensive, either requiring complex iterative optimization to find the optimal solution, or constructing the deraining framework in multiple stages with recurrent or progressive refinement steps involved.



In this work, we approach the single-image deraining problem from a different angle, aiming to propose a efficient while general derain method. First, our proposed method is model-free, which makes no assumptions of how the rain is generated. As we show in the experimental section, rain model assumption is somewhat not mandatory for achieving high-performance deraining, and sometimes such assumptions can even impair the deraining performance. Second, our proposed method follows a single-stage and does not require either iterative optimization or progressive refinement, thus leading to high-efficiency deraining. Overall, our main contributions are three-fold:

\begin{itemize}
   \item We propose the \textit{pixel-wise dilation filtering} for efficient deraining. A rainy image is filtered by the pixel-wise kernels estimated from a \textit{kernel prediction network} that can automatically and efficiently predict the suitable multi-scale kernels for each pixel. 
    
    \item To bridge the gap between synthetic and real data, we further propose the \textit{RainMix} component for simple yet effective data augmentation, which enables us to train networks for handling real rainy images
    
    \item We demonstrate the advantage of our method on synthetic and real-world rainy datasets, achieving both high-performance and high-efficiency deraining. As shown in Fig.~\ref{fig:teaser}, our method (\ie, EfDeRain) runs around $88$ times faster than the state-of-the-art method RCDNet \cite{rcd} with similar performance in terms of PSNR and SSIM. Furthermore, equipped with the \textit{RainMix}, EfDeRain can achieve much better visualization result on the real rainy images than RCDNet. %
\end{itemize}

\section{Related Work}\label{sec:related}

Based on the input data types, existing deraining algorithms can be largely categorized into two groups: video-based deraining, \eg, \cite{kim2015video}, and single image deraining. In this section, we discuss several representative DL-based single-image deraining methods, which are most relevant to ours.

In \cite{rcd}, a rain convolutional dictionary network (RCD-Net) is proposed for single image deraining, where the rain shapes are encoded by exploiting the intrinsic convolutional dictionary learning mechanism, leveraging the proximal gradient technique as the optimization method to seek the optimal solution. RCD-Net is designed as multiple stages to iteratively solve the deraining problem.
In particular, the algorithm iterates over two sub-steps,  1) updating the rain map by convolving with the learned rain kernel, and 2) updating the background layer, which are achieved by the two sub-net (M-net and B-net) at each stage of the RCD-Net. 
In \cite{cvid}, a conditional variational image deraining (CVID) network is proposed for draining, which uses the conditional variational auto-encoder (CVAE) architecture as the backbone. As a variational method, multiple predictions of the derained images from the input rainy image can be generated by the learned decoder, and the final single-image output is obtained by integrating these predictions. The learning stage of the CVID network primarily involves jointly training the variational encoder and decoder, with a spatial density estimation (SDE) module connected for estimating the intermediate rain map, which is then used to produce the latent space representation of the input ground-truth (clean) and rainy image pairs.
In \cite{jorder}, a multi-task network to perform joint rain detection and removal (JORDER) is proposed to solve the inverse problem of single-image deraining. The multi-task network jointly learns three targets: the binary rain-streak map, rain streak layers, and the clean background that is the final derained prediction.  Contextual dilated networks are incorporated to extract regional contextual information so that the learned feature can be invariant to rain streaks. To further process rain streaks with various directions and shapes, a recurrent process is adopted to progressively remove the rain streaks in stages.
In \cite{rescan}, a recurrent squeeze-and-excitation context aggregation net (RESCAN) is designed for single-image deraining. Similarly, the entire procedure is also performed in multiple stages, with a memory enabled RNN architecture to aggregate useful information of earlier stages. Within each stage, rain streaks exhibit various directions and shapes, and can be regarded as the accumulation of multiple rain streak layers. Thus, by incorporating a squeeze-and-excitation block, different alpha values can be properly assigned to each rain streak layers (\ie, feature maps), according to the intensity and transparency levels. 
In \cite{ddn}, further attempts to tackle the single-image deraining problem are made by modifying the classic ResNet architecture to better handle the image regression task at hand. By reducing the mapping range from input to output, the deraining learning process becomes more accessible. To guide the model in focusing on the structure of rain in the images, a priori image domain knowledge is used to shift the model attention to the high-frequency details, thus removing the background interference. 
In \cite{spa}, a spatial attentive network (SPANet) based on a two-round four-directional RNN architecture is proposed, where three standard residual blocks are used to extract features, and four spatial attentive blocks are used to identify rain streaks progressively in four stages. Next, two residual blocks are further adopted to reconstruct a clean background by removing rain streaks via the learned negative residuals. 
%

\begin{figure}[t]
\centering
\includegraphics[width=1.0\columnwidth]{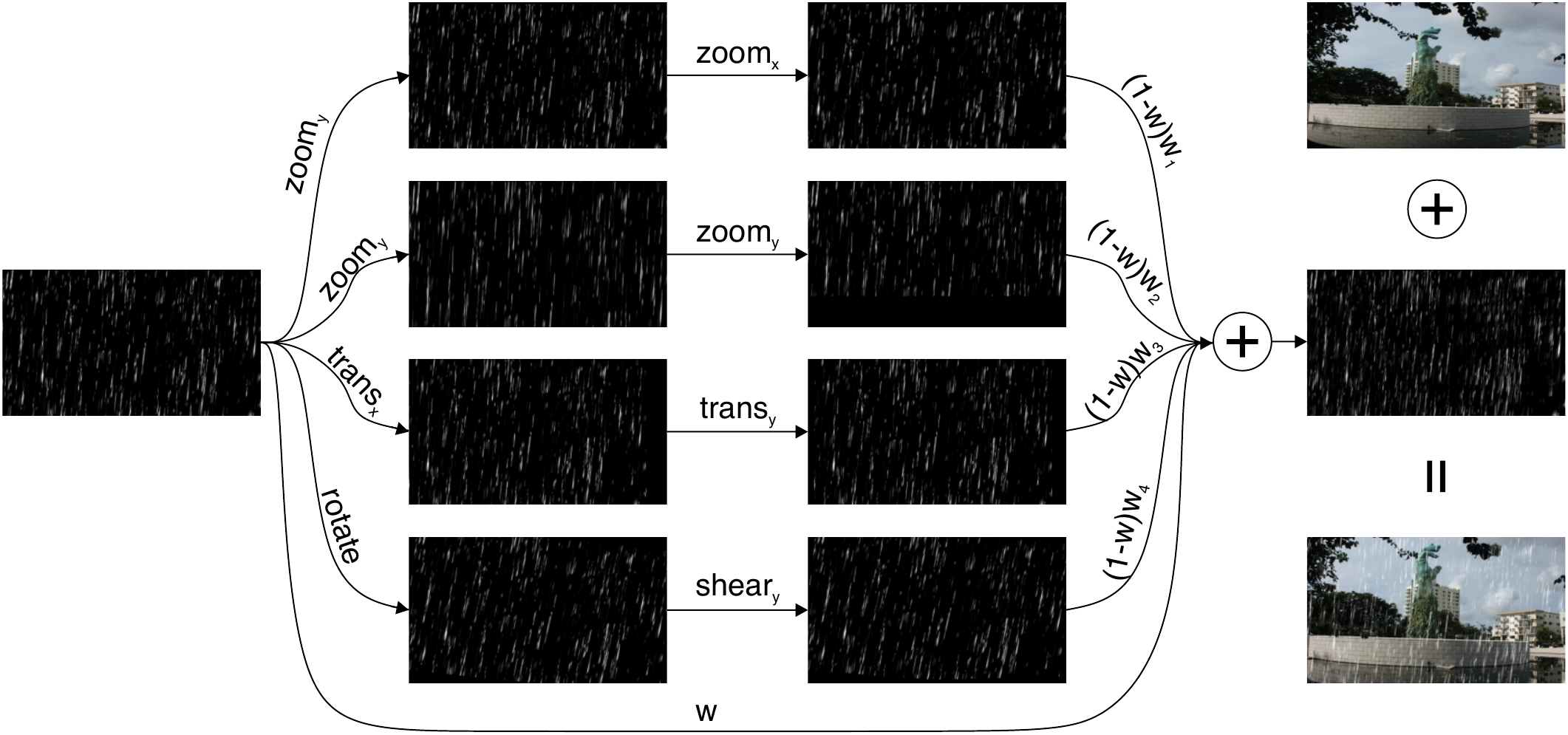}
\caption{An example of RainMix to generate a rainy image.}
\label{fig:rainmix}
\end{figure}
\begin{algorithm}[tb]
	{
		\caption{\small{Learning EfficientDeRain via RainMix}}\label{alg}
		\KwIn{$\text{KPN}(\cdot)$, fusion Convolution $\text{Conv}(\cdot)$, Loss function $\mathcal{L}$, Rainy Images $\mathcal{I}^{\mathrm{r}}$, Clean Images $\mathcal{I}$, real rain streak set $\mathcal{R}$, Operation set $\mathcal{O}=\{\text{rot},\text{shear}_{x/y},\text{trans}_{x/y}, \text{zoom}_{x/y}\}$.}
		\KwOut{Pre-trained Network $\text{KPN}(\cdot)$ and $\text{Conv}(\cdot)$.}
        \SetKwFunction{FMain}{RainMix}
        \SetKwProg{Fn}{Function}{:}{}
        \Fn{\FMain{$\mathcal{R}$}}{
        Sample a rain map $\mathbf{R}^\text{org}\sim\mathcal{R}$\;
        Initialize an empty map $\mathbf{R}^\text{mix}$\;
        Sample mixing weights $(w_1,w_2,w_3,w_4)\sim\text{Dirichlet}$\;
        \For{$i=1\ \mathrm{to}\ 4$}{
            Sample operations $(\text{o}_1,\text{o}_2,\text{o}_3)\sim\mathcal{O}$\;
            Combine via $\text{o}_{12}=\text{o}_2\text{o}_1$ and $\text{o}_{123}=\text{o}_3\text{o}_2\text{o}_1$\;
            Sample $\text{o}\sim\{\text{o}_{1},\text{o}_{12},\text{o}_{123}\}$\;
            $\mathbf{R}^\text{mix}+=w_i\text{o}(\mathbf{R}^\text{org})$
            }
        Sample a blending weight $w\sim\text{Beta}$\;
        \textbf{return} $\mathbf{R}=w\mathbf{R}^\text{org}+(1-w)\mathbf{R}^\text{mix}$\;
        }
        \textbf{End function}\;
 		\For{$i=1\ \mathrm{to}\ |\mathcal{I}^{\mathrm{r}}|$}{
 		    Generate rain map via $\mathbf{R}=\FuncSty{RainMix}(\mathcal{R})$\;
 		    Sample an image pair via $(\mathbf{I}^{\mathrm{r}},\mathbf{I})\sim(\mathcal{I}^{\mathrm{r}},\mathcal{I})$\;
 		    Sample $\mathbf{X}\sim\{\mathbf{I}^{\mathrm{r}},\mathbf{I}\}$ 
 		    and Perform $\mathbf{I}^{\text{rm}}=\mathbf{R}+\mathbf{X}$\;
            Predict pixel-wise kernels via $\mathbf{K}=\text{KPN}(\mathbf{I}^{\mathrm{rm}})$\;
            Derain via Eq.~\ref{eq:dilatedfilter} and get $\hat{\mathbf{I}}=\text{Conv}(\{\hat{\mathbf{I}}_l\})$\;
            Calculate Eq.~\ref{eq:loss} and do back-propagation\;
            Update parameters of $\text{KPN}(\cdot)$ and $\text{Conv}(\cdot)$\;
    	}
	}
\end{algorithm}

Overall, these existing methods share some commonalities. In particular, they often heavily rely on a presumed rain model to develop their algorithms such as \cite{jorder,rescan,rcd}, or require specific \textit{a priori} domain knowledge of the rain streaks such as \cite{spa,cvid,ddn}. Again, many of these methods are built on a recurrent or progressive framework, where later stage deraining can help refine the early stage results such as \cite{jorder,rescan,spa,rcd}. 
However, we believe that these strong prerequisites and constraints can potentially hinder both universality and efficiency of the derain method and its real-world deployment. In this work, we originally propose a deraining method from a different perspective, aiming to address both issues.

\section{Methodology}\label{sec:method}

\subsection{Pixel-wise Image Filtering for Deraining}\label{subsec:kernel_pred_}
In this part, we propose an image filtering-based method for model-free deraining.
In general, the rain can be regarded as a kind of degradation that may cause effects such as occlusion, fog, motion blur, \etc. As a result, it is reasonable to use image filtering methods to handle it, which can be efficient and effective for various degradations.
Specifically, we process an input rainy image $\mathbf{I}^{\text{r}}\in\mathds{R}^{H\times W}$ \footnote{\scriptsize We exemplify our method via the gray-scale image for better understanding. For color images, we can handle each channel independently.} via pixel-wise filtering
\begin{align}\label{eq:imagefilter}
\hat{\mathbf{I}} = \mathbf{K}\circledast\mathbf{I}^{\text{r}},
\end{align}
%
where $\hat{\mathbf{I}}\in\mathds{R}^{H\times W}$ is the estimated derained image, and $\circledast$ denotes the pixel-wise filtering operation where each pixel is processed by its exclusive kernel and $\mathbf{K}\in\mathds{R}^{H\times W \times K^2}$ contains the kernels of all pixels.
Specifically, when deraining the $\mathbf{p}$-th pixel of $\mathbf{I}^{\text{r}}$, we have its exclusive kernel, \ie, the vector at $\mathbf{p}$-th position of $\mathbf{K}$, which is reshaped and denoted as  $\mathbf{K}_p\in\mathds{R}^{K\times K}$. We use $\mathbf{p}$ as 2D coordinates for a pixel.
Then, we can predict the derained pixel by
%
\begin{align}\label{eq:pixelfilter}
\hat{\mathbf{I}}(\mathbf{p}) = \sum_{\mathbf{t},\mathbf{q}=\mathbf{p}+\mathbf{t}}\mathbf{K}_{p}(\mathbf{t})\mathbf{I}^{\text{r}}(\mathbf{q}),
\end{align}
%
where $\mathbf{t}$ ranges from $(-\frac{K-1}{2}, -\frac{K-1}{2})$ to $(\frac{K-1}{2}, \frac{K-1}{2})$.

To realize effective deraining with the simple pixel-wise filtering, we should consider the following \textit{challenges:} \ding{182} \textit{how to estimate spatial-variant, scale-variant, and semantic-aware kernels effectively and efficiently.} Rain may cause streak occlusion, fog, and blur, at different positions with different appearances. 
For example, rain streaks could exhibit different scales, 
directions, and transparency across the image and semantically related to the image contents, \eg, the scene depth \cite{hu2019cvpr}.
As a result, the pixel-wise kernels should be adapted to scene information, the spatial and scale variations of rain streaks. 
Obviously, hand-craft designed kernels can hardly satisfy such requirements. 
To address this challenge, we propose the \textit{multi-dilated-kernel prediction network} in Sec.~\ref{subsec:multi_dilated_kernel pred}, which takes the rainy image as the input and predict multi-scale kernels for each pixel via the deep neural network (DNN).
\ding{183} \textit{how to train a powerful deraining DNN to bridge the gap to real rain with synthetic data.} Most of the existing deraining DNNs are trained on the synthetic data. However, there is still a gap between real and synthetic rain. Bridging this gap is of great importance for real-world applications. We propose a simple yet effective rain augmentation method, denoted as \textit{RainMix} in Sec.~\ref{subsec:rainmix} to reduce such a gap.

\begin{figure}[t]
	\centering
	\includegraphics[width=1.0\columnwidth]{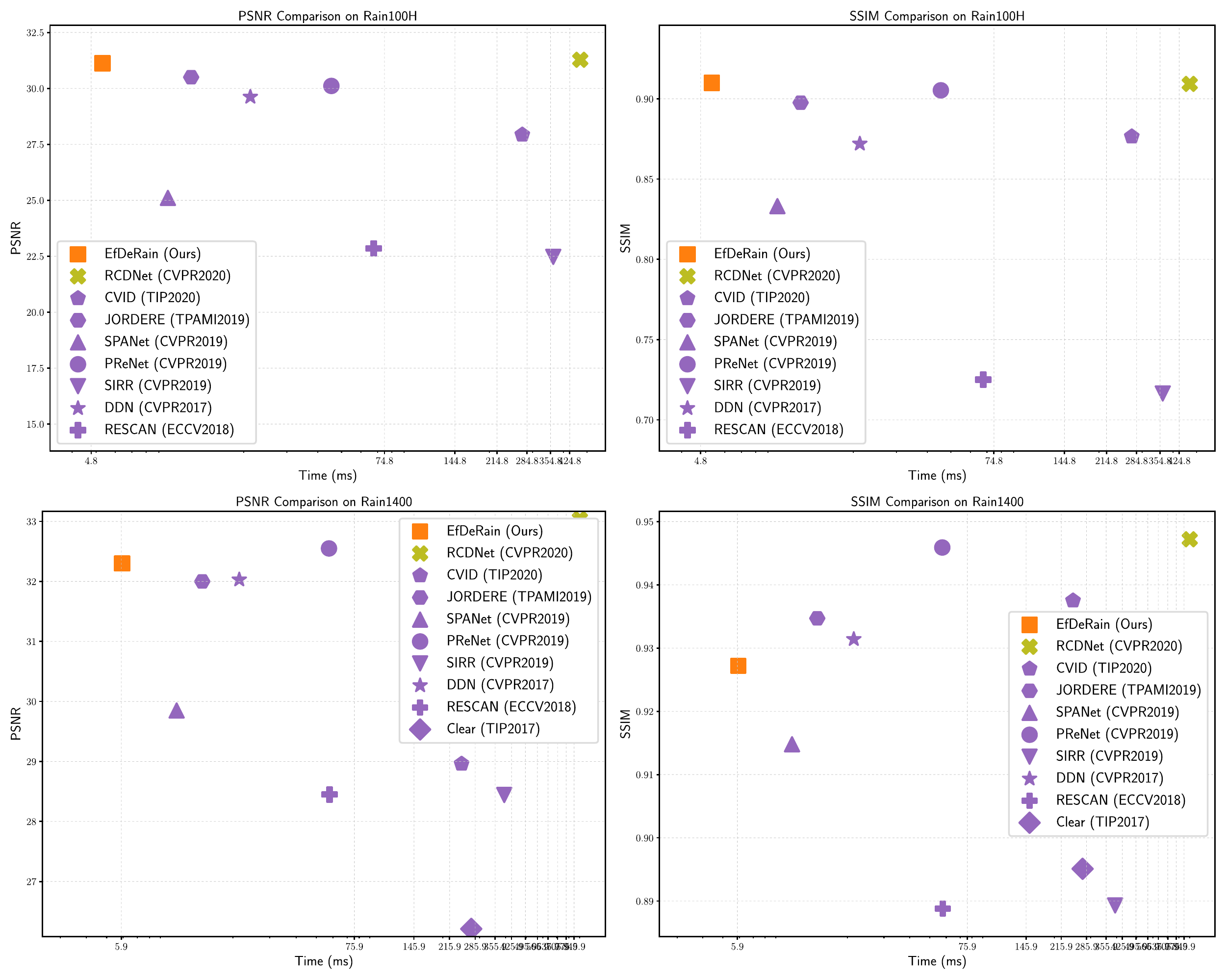}
	\caption{Comparing our method EfDeRain with nine baseline methods on Rain100H \cite{yang2017cvpr,jorder} and Rain1400 \cite{ddn}  with PSNR vs. Time and SSIM vs. Time plots.}
	\label{fig:cmp1}
\end{figure}
\begin{figure}[t]
	\centering
	\includegraphics[width=1.0\columnwidth]{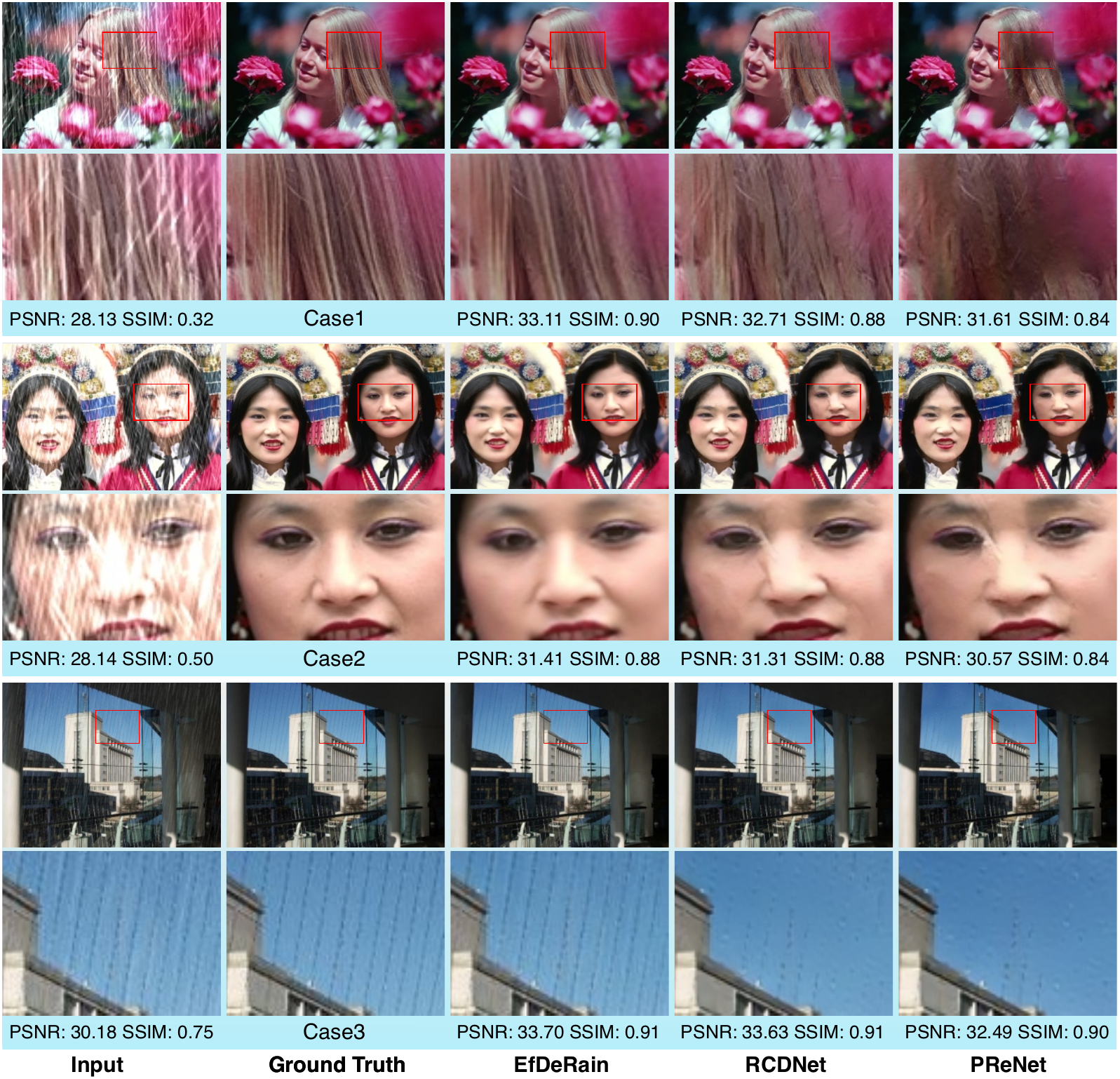}
	\caption{Three visualization results of EfDeRain, RCDNet \cite{rcd}, and PReNet \cite{ren2019cvpr} on Rain100H (Case1 and Case2) and Rain1400 (Case3). We magnify the main difference regions to highlight the advantages of our method.}
	\label{fig:cmp_rain100_vis}
\end{figure}

\subsection{Learnable Pixel-wise Dilation Filtering}\label{subsec:multi_dilated_kernel pred}

\subsubsection{Kernel prediction network.}
Inspired by recent works on image denoising \cite{BakoACMTOG2017,MildenhallCVPR2018}, we propose to estimate the pixel-wise kernels $\mathbf{K}$ for deraining by taking the rainy image as input,
%
\begin{align}\label{eq:kpn}
\mathbf{K} = \text{KPN}(\mathbf{I}^\text{r}),
\end{align}
where $\text{KPN}(\cdot)$ denotes a UNet-like network and we show the architecture in Fig.~\ref{fig:pipeline}. 
By offline training on rainy-clean image pairs, the kernel prediction network can predict spatial-variant kernels that adapt to the rain streaks with different thickness and strength while preserving the object boundary. We show a deraining example in Fig.~\ref{fig:pipeline}, where we validate our method on six representative regions and observe that: \ding{182} Our method can effectively remove the rain steak while recovering the occluded boundary, as shown in the region one (R1) (see Fig.~\ref{fig:pipeline} (c)). \ding{183} The predicted kernels can adapt to the rain with different strengths. As shown in Fig.~\ref{fig:pipeline} (c), from R2 to R5, the rain strength gradually becomes weak and our method can remove all trace effectively.
Moreover, according to the visualization of predicted kernels, our network can perceive the positions of rain streak. As a result, the predict kernels assign higher weights to non-rainy pixels and lowers ones to rainy pixels, confirming the effectiveness of our method. \ding{184} According to the results on R6, our method does not harm the original boundary and makes it even sharper.

\subsubsection{Multi-dilated image filtering and fusion.}
To enable our method to handle multi-scale rain streaks without harming the efficiency, we extend each predicted kernel to three scales via the idea of dilated convolution \cite{Yu2016iclr}.

Intuitively, when the rain streak covers a large region of the image, the large scale kernel can use related pixels far from the rain for effective deraining.
A simple solution is to predict multi-scale kernels directly, leading to extra parameters and time costs.
Alternatively, we propose to extend the pixel-wise filtering in Eq.~\ref{eq:imagefilter} to pixel-wise dilation filtering \cite{Yu2016iclr} for convolution layer,
%
\begin{align}\label{eq:dilatedfilter}
\hat{\mathbf{I}}_l(\mathbf{p}) = \sum_{\mathbf{t},\mathbf{q}=\mathbf{p}+l\mathbf{t}}\mathbf{K}_{p}(\mathbf{t})\mathbf{I}^{\text{r}}(\mathbf{q}),
\end{align}
%
where $l$ is the dilation factor to control the applying range of the same filter. In practical, we consider four scales, \ie, $l=1,2,3,4$. With Eq.~\ref{eq:dilatedfilter}, we can get four derained images, \ie, $\hat{\mathbf{I}}_1$, $\hat{\mathbf{I}}_2$, $\hat{\mathbf{I}}_3$, and $\hat{\mathbf{I}}_4$. Then, we use a convolution layer with the size $3\times3$ to fuse the four images and obtain the final output.

\subsubsection{Loss function.}
We consider two loss functions for training the network, \ie, $L_1$ and SSIM loss functions. Given the derained image, \ie, $\hat{\mathbf{I}}$, and the clean image $\mathbf{I}$ as ground truth, we have
%
\begin{align}\label{eq:loss}
\mathcal{L}(\hat{\mathbf{I}}, \mathbf{I}) = \|\hat{\mathbf{I}}-\mathbf{I}\|_1-\lambda~\text{SSIM}(\hat{\mathbf{I}}, \mathbf{I})
\end{align}
%
where we fix $\lambda=0.2$ for all experiments.

\subsection{RainMix: Bridging the Gap to Real Rain}\label{subsec:rainmix}

How to reduce the gap of synthetic rainy images to real data is still an open problem.
%
%
In this section, we explore a novel solution \textit{RainMix} to address this challenge. 
%
%
\citet{garg2006siggraph} conducted a detailed study about the appearance of rain streaks and constructed a dataset of real rain streaks by considering different lighting and viewing conditions.
Even then, it is still hard to say the collected real rain is exhaustive, which covers all possible situations in the real world, since the rain has diverse appearances due to the influences of various natural factors, such as the wind, the light reflection, and refraction, \etc.
However, it is reasonable to use these real rain streaks to generate more rain streaks through transformations that may occur in the real world, \eg, zooming, shearing, translation, and rotation. \textit{RainMix} is originally designed based on this such intuition.

We show our RainMix-based learning algorithm in Algorithm~\ref{alg}. At each training iteration, we generate a rain map via \textit{RainMix} and add it to the clean or rainy images, to obtain a new rain image for training the kernel prediction network and the fusion convolution layer. Our \textit{RainMix} randomly samples a rain map from the dataset of real rain streaks \cite{garg2006siggraph} (\ie, line 2 in Algorithm~\ref{alg}) and performs three transformations on the rain map through randomly sampled and combined operations (\ie, line 5-9 in Algorithm~\ref{alg}). Finally, the three transformed rain maps are aggregated via the weights from Dirichlet distribution and further blended with the original sampled rain map via the weight from Beta distribution (\ie, line 4 and 11 in Algorithm~\ref{alg}). Intuitively, the multiple random processes simulate the diverse rain appearance patterns in the real world. We give an example of RainMix for generating a rainy image in Figure~\ref{fig:rainmix}.

\begin{figure}[t]
	\centering
	\includegraphics[width=1.0\columnwidth]{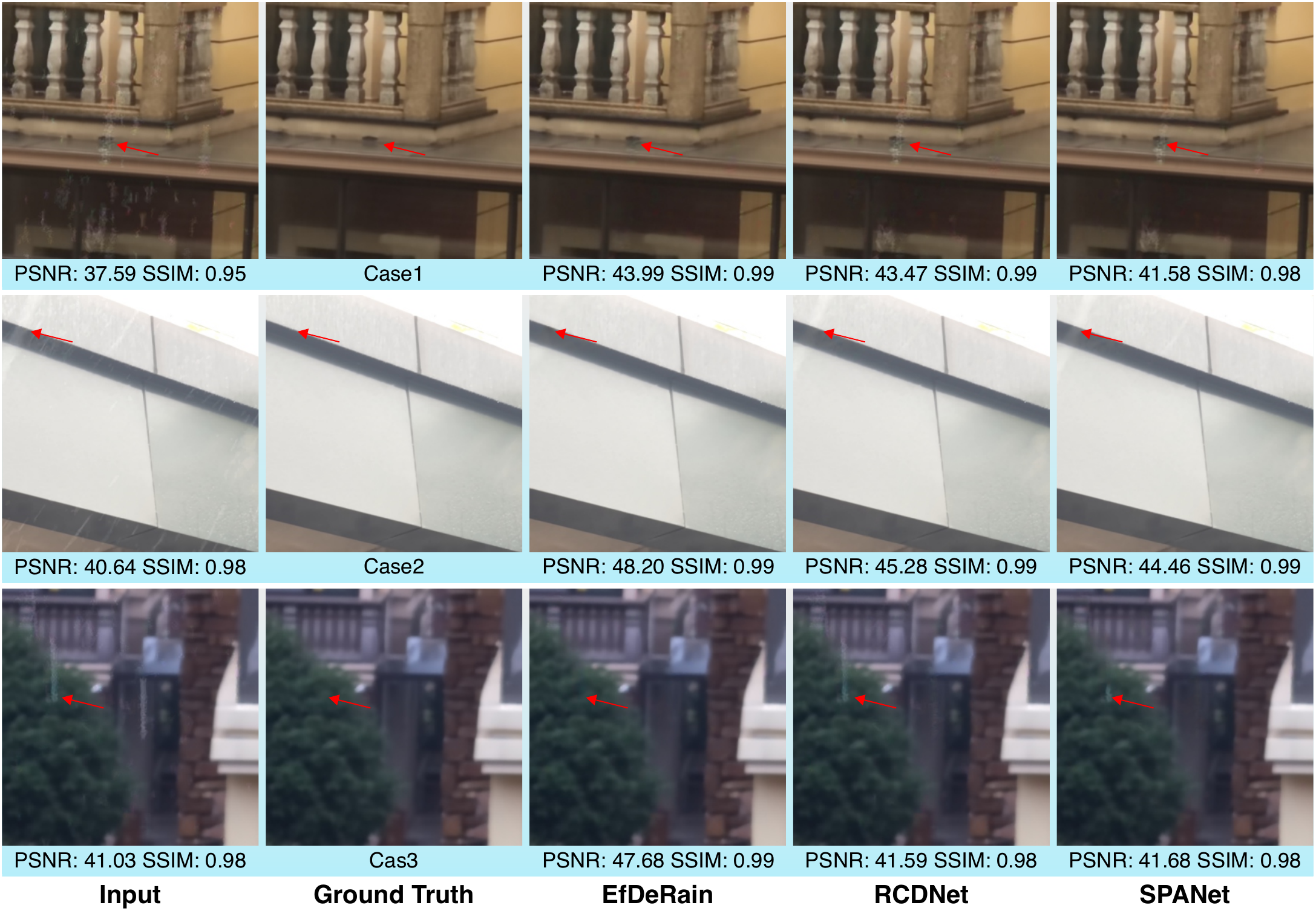}
	\caption{Three visualization results of EfDeRain, RCDNet \cite{rcd} and SPANet \cite{spa} on the real-world SPA dataset. The \textcolor{red}{red arrow} shows the main difference between EfDeRain and other two methods. }
	\label{fig:cmp_spa_vis}
\end{figure}
\begin{figure}[t]
	\centering
	\includegraphics[width=1.0\columnwidth]{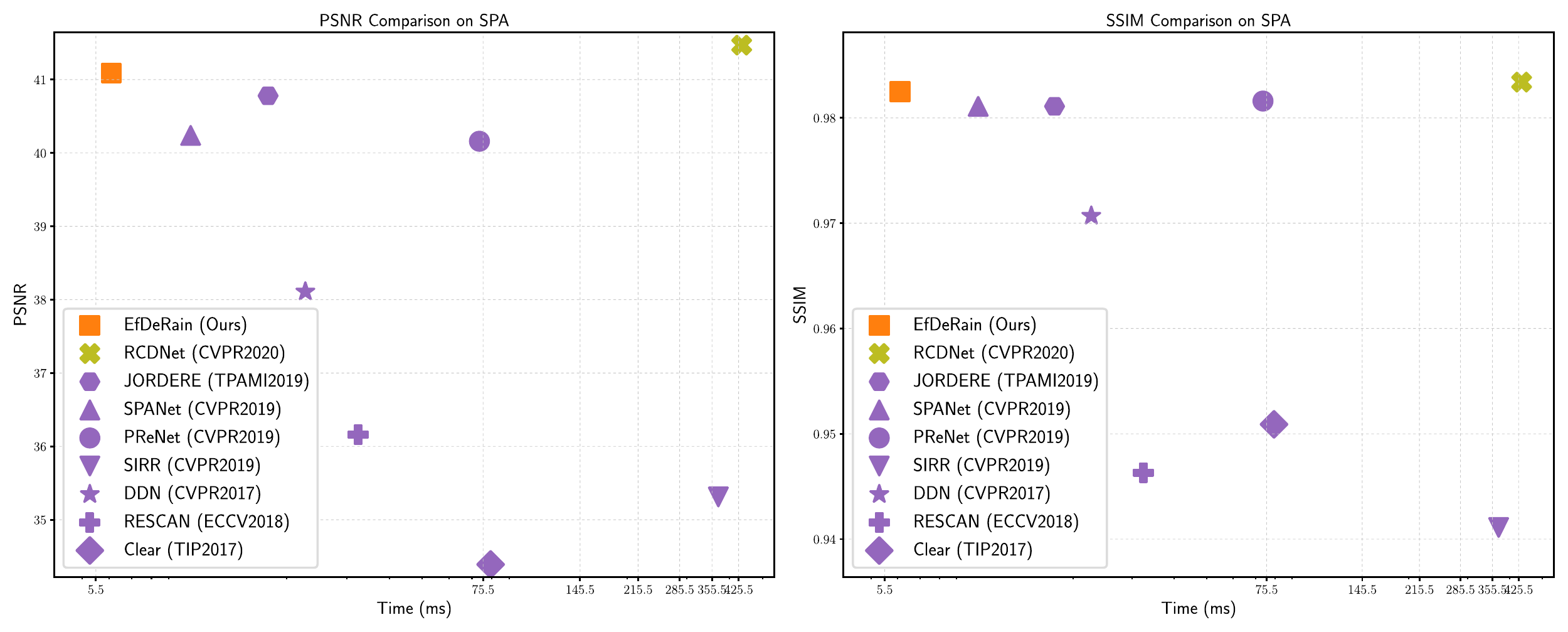}
	\caption{Comparing our method EfDeRain with eight baseline methods on the real SPA dataset \cite{spa} via PSNR vs. Time and SSIM vs. Time plots. The CVID method is not evaluated on SPA, thus not included in the plots.}
	\label{fig:cmp2}
\end{figure}
\begin{figure*}[th!]
	\centering
	\includegraphics[width=2.1\columnwidth]{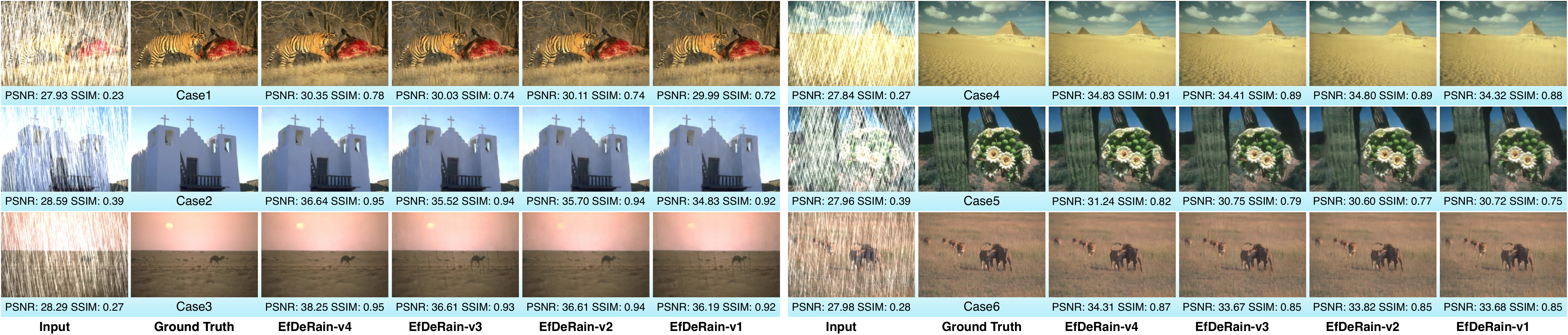}
	\caption{Six visualization results of four variants of EfDeRain: v1 denotes the pixel-wise filtering based on KPN without dilation structure; v2 is the pixel-wise dilation filtering for deraining. v1 and v2 are trained based on the $L_1$ loss. v3 and v4 have the same structure with v2 but are trained via $L_1$ and SSIM loss functions, \ie, Eq~\ref{eq:loss}. In addition, the v4 uses RainMix for training.}
	\label{fig:ablation_vis}
\end{figure*}

\section{Experiments}\label{sec:exp}

\subsection{Setups}

\subsubsection{Datasets.}  To comprehensively validate and evaluate our method, we conduct the comparison and analysis experiments on 4 popular datasets, including \emph{Rain100H} \cite{yang2017cvpr,jorder}, \emph{Rain1400} \cite{ddn} synthetic dataset, the recent proposed \emph{SPA} real rain dataset \cite{spa}, and the real \emph{Raindrop} dataset \cite{qian2018cvpr}.

\subsubsection{Metrics.} We employ the commonly used peak signal to noise ratio (PSNR) and structural similarity (SSIM) as the quantitative evaluation metric for all datasets. In general, the larger PSNR and SSIM indicate better deraining results.

\subsubsection{Baselines.}
To be comprehensive, we perform a large-scale evaluation, to compare with a total of 14 (=9+5) state-of-the-art deraining methods, \ie, 9 baselines for the derain streak task (removing rain streak), including rain convolutional dictionary network (RCDNet) \cite{rcd}, conditional variational image deraining (CVID) \cite{cvid}, 
joint rain detection and removing (JORDERE) \cite{jorder},
spatial attentive deraining method (SPANet) \cite{spa}, progressive image deraining network (PReNet) \cite{ren2019cvpr},
semi-supervised transfer learning for rain removal (SIRR) \cite{wei2019cvpr},
recurrent squeeze-and-excitation context aggregation net (RESCAN) \cite{rescan}, 
deep detail network \cite{ddn}, and Clear \cite{fu2017tip}. Furthermore, for deraindrop task (\ie, removing train drop) on the Raindrop dataset \cite{qian2018cvpr}, we compare another 5 state-of-the-art methods \cite{li2019cvpr} as baselines, including GMM \cite{li2016cvpr}, JORDER \cite{yang2017cvpr}, DDN \cite{ddn}, CGAN \cite{zhang2019tcsvt}, DID-MDN \cite{zhang2018iccv}, and DeRaindrop \cite{qian2018cvpr}.
Note that, the time cost of all compared method are one-by-one evaluated on the same PC with the Intel Xeon CPU (E5-1650) and NVIDIA Quadro P6000 GPU.

\subsection{Comparison on Rain100H\&1400 Dataset}

We compare our method with the 9 baseline methods in Fig.~\ref{fig:cmp1} on Rain100H and Rain1400 datasets. In general, our method, \ie, EfDeRain, achieves the lowest time cost while obtaining comparable PSNR or SSIM with the top methods on both datasets. More specifically, on the challenging Rain100H dataset where the rain streaks cover a large portion of the image, we observe that EfDeRain achieves almost the same PSNR and SSIM with the rank$1$ method, \ie, RCDNet \cite{rcd}, while being $88$ times faster. Compared with other methods, \eg, JORDERE \cite{jorder} and CVID \cite{cvid}, our method exhibits clear advantages on both recovery quality and efficiency. For example, EfDeRain achieves 11. 4\% relative PSNR improvement over CVID while running over 50 times faster. Similarly, in terms of the Rain1400 dataset, our method still achieves the lowest time cost while maintaining the competitive PSNR and SSIM in line with the state-of-the-art methods, \eg, PReNet and RCDNet. The main reason is likely to be that the rain streak of Rain1400 is slighter than Rain100H and a lot of the regions are not covered by the rain. Meanwhile, as shown in the Fig.~\ref{fig:pipeline}, our method can not only remove rain streaks but enhance the object boundary, leading to the negative score of PSNR and SSIM that evaluate the recovery quality instead of the enhancement capability.

We further compare the visualization results of EfDeRain with the state-of-the-art baseline methods, \ie, RCDNet and PReNet, in Fig.~\ref{fig:cmp_rain100_vis} and observe that: \ding{182} In Case2, EfDeRain can remove the rain streaks more effectively than the other two methods since there are clear rain traces on derained images, \ie, the white streaks near the nose. \ding{183} Compared with the baseline methods, EfDeRain removes the rain streak while recovering the original details well. In Case1 and Case3, RCDNet and PReNet remove or destroy the original streak-like details, \eg, the girl's hair in Case1 and the fine lines in Case3. In contrast, our method preserves these details while removing the rain effectively, demonstrating that our method could understand the scene better and predict kernels for different pixels.

\subsection{Comparison on Real-world SPA Rain Dataset}

We further compare our method with 8 baseline methods on the SPA dataset \cite{spa}, where the rainy image is real and its ground truth is obtained by human labeling and multi-frame fusion. As shown in Fig.~\ref{fig:cmp2}, our method achieves almost the same PSNR and SSIM with the top method, \ie, RCDNet, and outperforms all other baselines while running over 71 times more efficient than RCDNet.

We also visually compare our method with RCDNet and SPANet in Fig.~\ref{fig:cmp_spa_vis}. Obviously, all the results demonstrate that our method can handle various rain traces with different patterns and achieves better visualization results than RCDNet and SPANet. In particular, both RCDNet and SPANet fail to remove the wider rain streak in Case2 while our method successfully handles all rain streaks and the obtained derained image is almost the same with the ground truth.

\subsection{Comparison on Real-world Raindrop Dataset}
%
\begin{table}[t]
	\centering
	\small
	\resizebox{0.99\linewidth}{!}{
	\begin{tabular}{c|ccccccc}
		\toprule
		\rowcolor{tabgray}    &  GMM & JORDER & DDN & CGAN & DID-MDN & DeRaindrop & EfDeRaiin \\ \midrule
		PSNR &  24.58  &  \cellcolor{top3} 27.52   &   25.23    &   21.35   &  24.76  & \cellcolor{top1} 31.57  & \cellcolor{top2} 28.48 \\ 
		SSIM &   0.7808  &  \cellcolor{top3} 0.8239  &    0.8366   &    0.7306  &   0.7930   &  \cellcolor{top1} 0.9023 & \cellcolor{top2} 0.8971 \\ \bottomrule
	\end{tabular}
	}
	\caption{Experimental results on Raindrop dataset \cite{qian2018cvpr}. The baseline results are reported by \cite{li2019cvpr} and \cite{qian2018cvpr}. We highlight the top three results with red, yellow, and green, respectively.}	
\end{table}

Besides the rainy streak image dataset, we also compare our method on the deraindrop task to show the generalization capability of our method. We train our network on the Raindrop dataset \cite{qian2018cvpr} and compare it with 6 state-of-the-art baseline methods. In particular, the method DeRaindrop \cite{qian2018cvpr} is specifically designed for this problem, where the region of the raindrop is perceived by an attentive-recurrent network. Our method without changing any architectures or hyper-parameters, achieves the second-best results with competitive SSIM to DeRaindrop, and outperforms all other deraining methods, demonstrating both the effectiveness and generality of our method.

\subsection{Ablation Study}\label{subsec:ablation}

\begin{figure}
	\centering
	\includegraphics[width=1.0\columnwidth]{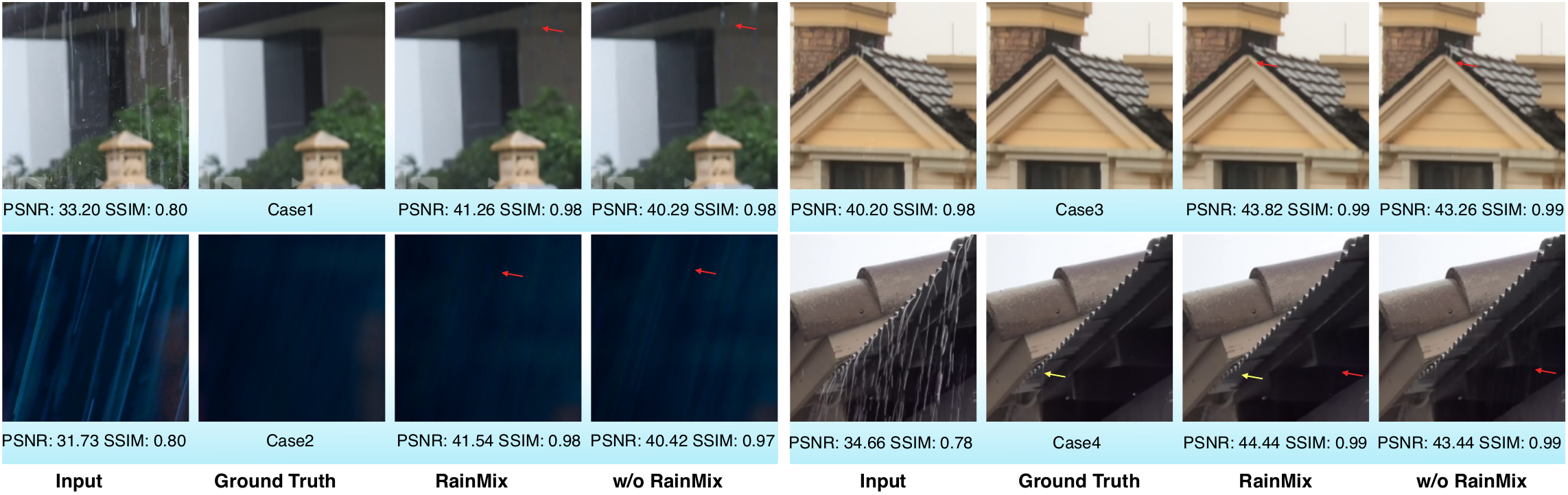}
	\caption{Four visualization results of our EfDeRain with or without RainMix for training. The \textcolor{red}{red arrow} shows the main difference between our two versions. The \textcolor{darkorange}{yellow arrow} shows that the rain drops not labeled by human are also removed by our method. }
	\label{fig:rainmix_vis}
\end{figure}
\begin{table}[t]
	\centering
	\small
	\resizebox{1.0\linewidth}{!}{
		\begin{tabular}{l|cccc}
			\toprule
			\rowcolor{tabgray}  EfDeRain & w/o dilation(v1)  &    +dilation(v2)  &  +ssim loss(v3)  &  +RainMix(v4)    \\ \midrule
		    PSNR     &   \cellcolor{top1-5}    30.12      &     \cellcolor{top1-4}  30.27       &         \cellcolor{top1-3} 30.35      &      \cellcolor{top1-2} 31.02       \\ 
			SSIM     &    \cellcolor{top1-5}   0.8834       &   \cellcolor{top1-4}    0.8918        &      \cellcolor{top1-3}    0.8970      &       \cellcolor{top1-2}  0.9079      \\
			Time~(ms)     &      4.41       &      6.05      &         5.93      &       5.97      \\
			\bottomrule
		\end{tabular}
	}
	\caption{Ablation study on Rain100H. We consider four variants of EfDeRain: v1 denotes the pixel-wise filtering based on KPN without dilation structure;  v2 is the pixel-wise dilation filtering for deraining. v1 and v2 are trained based on the $L_1$ loss. v3 and v4 have the same structure with v2 but are trained via $L_1$ and SSIM loss functions, \ie, Eq~\ref{eq:loss}. In addition, the v4 uses RainMix for training.}
	\label{tab:ablation}
\end{table}

In this subsection, we validate the effectiveness of our main contributions, \ie, pixel-wise dilation filtering~(Sec.~\ref{subsec:multi_dilated_kernel pred}) and RainMix~(Sec.~\ref{subsec:rainmix}), on Rain100H dataset. We also discuss the effectiveness of the loss function in Eq.~\ref{eq:loss}. More specifically, we develop four variants of our method: the first version (\emph{v1}) is the pixel-wise filtering based on KPN without the dilation structure;  the second version (\emph{v2}, \ie, Sec.~\ref{subsec:multi_dilated_kernel pred}) denotes the pixel-wise dilation filtering for deraining. \emph{v1} and \emph{v2} are trained based on the $L_1$ loss. \emph{v3} and \emph{v4} share the same structure with \emph{v2} but are trained via $L_1$ and SSIM loss functions, \ie, Eq~\ref{eq:loss}. In addition, the final version (\emph{v4}) uses RainMix in Sec.~\ref{subsec:rainmix} for training.

As shown in Table~\ref{tab:ablation}, the PSNR and SSIM of the 4 versions gradually increase and reach the highest performance on our final version with the pixel-wise dilation filtering trained via the RainMix as well as $L_1$ and SSIM loss functions. This demonstrates that all of our main contributions are beneficial for effective deraining. Moreover, we also analyze the time cost of the four variants and observe that: the proposed dilation filtering only leads to around $1.5$ms additional costs.As shown in Table~\ref{tab:ablation}, the PSNR and SSIM of the four versions gradually increase and reach the highest performance on our final version with the pixel-wise dilation filtering trained via the RainMix as well as $L_1$ and SSIM loss functions, demonstrating that all of our main contributions benefit to the effective deraining. Moreover, we also show the time cost of four variants and observe that: the proposed dilation filtering only leads to around $1.5$ms more costs. 

We further validate the advantages of our contributions through the Rain100H visualization results in Fig.~\ref{fig:ablation_vis} and observe that: \ding{182} In general, our final version can not only remove the heavy rain streak but recovering the original details, thus achieving the highest PSNR and SSIM scores. \ding{183} When comparing the EfDeRain-\emph{v1} with \emph{v2} (\ie, dilation-enhanced \emph{v1}), we can conclude that the dilation structure clearly facilitates to remove more rain streaks. For example, in Case~1, 2 and 3, the rain traces in \emph{v1} have been obviously suppressed in \emph{v2}. \ding{184} The SSIM loss function helps recover the details but enhances the rain streak. For example, in Case3, with the SSIM loss function(\ie, EfDeRain-\emph{v3}), the boundary of the sun in \emph{v3} becomes much sharper than that in \emph{v2} and \emph{v1}. However, the rain streak boundary becomes obvious as well. We observe similar results in other cases. \ding{185} By combing dilation structure, SSIM loss function, and RainMix, our final version, \ie, EfDeRain-\emph{v4}, is able to remove the heavy rain effectively while recovering the details very well.
Moreover, we conduct the visualization comparison on the SPA dataset in Fig.~\ref{fig:rainmix_vis} to validate the effectiveness of our RainMix. As a result, our RainMix enhances the capability of our method in removing the real rain trace even through the rain patterns are quite diverse. In all cases, if we do not employ the RainMix, we observe that there could always exist some heavy rain traces that can not be addressed, which are indicated by the red arrow in the figure.

\section{Conclusions}\label{sec:concl}

In this paper, we propose a novel model-free deraining method denoted as \emph{EfficientDeRain}. Our method can not only achieve the significantly high performance but runs over 80 times more efficient than the state-of-the-art method. Two major contributions benefit to our final performance: \textit{First}, we proposed and designed the novel \textit{pixel-wise dilation filtering} where each pixel is filtered by multi-scale kernels estimated from an offline trained kernel prediction network. \textit{Second}, we propose a simple yet effective data augmentation method for training the network, \ie, \textit{RainMix}, bridging the gap between synthesis data and real data. Finally, we perform a large-scale evaluation to comprehensively validate our method on popular and challenging synthesis datasets, \ie, Rain100H and Rain1400, and real-world datasets, \ie, SPA and Raindrop, all of which demonstrate the advantage of our method in terms of both efficiency and deraining quality.

In the future, we will study the effects of deraining to other computer vision tasks, \eg., object segmentation \cite{guo2018frequency,guo2017frequency} and object tracking \cite{guo2020spark,guo2020selective,guo2017learning,guo2017structure}, with the state-of-the-art DNN testing works \cite{issta19_deephunter,du2019deepstellar,xie2019diffchaser,ase18_gauge,issre18_mutation,saner19_deepct}. We also would like to study the single-image deraining from the view of adversarial attack methods, \eg, \cite{guo2020abba,wang2020amora,cheng2020pasadena,arxiv20_cosal,arxiv20_retinopathy,arxiv20_xray}.

\footnotesize
\bibliography{ref}

\end{document}